\documentclass[conference]{IEEEtran}
\IEEEoverridecommandlockouts

\usepackage{cite}
\usepackage{amsmath,amssymb,amsfonts}
\usepackage{algorithmic}
\usepackage[ruled,vlined]{algorithm2e}
\usepackage{graphicx}
\usepackage[table]{xcolor}
\usepackage{multirow}
\usepackage{subfigure}
\usepackage[nameinlink,noabbrev]{cleveref}
\crefformat{equation}{(#2#1#3)}
\Crefformat{equation}{(#2#1#3)}
\usepackage{flushend}
\usepackage{booktabs}
\usepackage{array}
\usepackage{xcolor,soul}
 
\DeclareMathOperator*{\argmin}{argmin}

\usepackage{tikz}
\usetikzlibrary{shapes.geometric}

\newcommand{\maybeincludegraphics}[2][]{\IfFileExists{#2}{\includegraphics[#1]{#2}}{\fbox{\parbox[c][1.4in][c]{0.9\linewidth}{\centering Missing figure: #2}}}}

\def\BibTeX{{\rm B\kern-.05em{\sc i\kern-.025em b}\kern-.08em
    T\kern-.1667em\lower.7ex\hbox{E}\kern-.125emX}}

\begin{document}

\title{Explainable Runtime Dependency Tracking for AI-RAN Conflict Monitoring}

\author{
	\IEEEauthorblockN{Christie Djidjev}
	\IEEEauthorblockA{\textit{Idaho National Laboratory} \\
		Idaho Falls, United States \\
		Christie.Djidjev@inl.gov}
	\and
	\IEEEauthorblockN{Nicholas Kaminski}
	\IEEEauthorblockA{\textit{Idaho National Laboratory} \\
		Idaho Falls, United States\\
		Nicholas.Kaminski@inl.gov}	
}

\maketitle

\begin{abstract}
Future AI-integrated Radio Access Networks (AI-RAN) will combine open programmability with learning-enabled xApps, rApps, and control functions that act on shared parameters and key performance indicators (KPIs).
For conflict monitoring, it is not enough to know which applications are deployed; the system must also know whether the parameter--KPI dependencies assumed by runtime diagnosis remain valid under the current operating regime.
This paper studies a lightweight monitoring primitive for that purpose: tracking an interpretable dependency representation from streaming telemetry events.

We represent active dependencies by a Boolean matrix and use Boolean matrix multiplication to check whether recent parameter-activity and KPI-response events are consistent with the current estimate.
We propose a sliding-window inference procedure that reuses the estimate when it remains consistent and recomputes it when recent observations indicate structural change.
The tracker is intended as an explainable signal for conflict diagnosis and slow-loop model refresh, not as an autonomous mitigation mechanism.
Experiments on controlled Boolean event streams show efficient and accurate tracking under dependency changes and Boolean observation noise.
\end{abstract}


\section{Introduction}

Radio access networks are moving toward open, programmable, and increasingly AI-integrated architectures.
This transition is a clear priority for both commercial and defense communities.
Open Radio Access Network (O-RAN) exposes control interfaces to enable independently developed third-party applications, termed xApps and rApps.
These applications optimize mobility, slicing, load balancing, energy consumption, and resource allocation \cite{abdalla_oran,polese_coloran}.
The OUSD(R\&E) FutureG Office has identified programmable, open-interface RANs as foundational infrastructure for future defense communications.
This office actively funds open-source RAN stacks, security-focused xApp and rApp development, and resilient tactical deployments under its Spectrum Forward initiative \cite{futureG2024sbir,pimentel2024dod}.
The FutureG-backed OCUDU initiative is producing an open, carrier-grade, AI-native RAN software stack for both commercial and mission-critical defense operations \cite{linux2026ocudu,deepsig2025ocudu}.
AI-RAN further develops learning, prediction, and adaptive decision-making as core parts of the control loop \cite{montebugnoli_clo5er,lindra_airan}.
Future networks trend toward multiple adaptive control functions that change with traffic, context, policy, and the operating environment.

This flexibility creates a runtime reasoning problem that is especially acute in military scenarios.
Control functions operating in the same RAN may interact through shared parameters, shared KPIs, or feedback through observed network state.
These interactions allow conflicts that erode the ability to achieve intended objectives.
In a commercial setting, conflicts may degrade throughput or energy efficiency.
In a tactical setting, undetected control-function conflicts can degrade mission-critical links, mask adversarial interference signatures, or prevent adaptive responses from taking effect.
Prior FutureG SBIR solicitations have specifically identified xApps and rApps as a priority for achieving the agility and resilience necessary to support Warfighter communications \cite{futureG2024sbir}.

Some conflicts may be visible from application subscriptions or control objectives.
Others arise only through the propagation of parameter changes to KPIs and then subsequent control decisions.
These indirect interactions are difficult to identify in advance because the relevant parameter--KPI relationships may be hidden, nonlinear, or operating-regime dependent.
In a tactical context, regime changes may be abrupt; node loss, jamming onset, or dynamic insertion of new control functions can shift the parameter--KPI relationships within seconds.
As highlighted in FutureG FRONTIER workshop, adversarial actions further complicate tactical environments with dependency shifts that conventional monitoring cannot anticipate \cite{futureG2024frontier}.
Runtime tracking must therefore ask whether dependencies used by monitoring logic are consistent with recent telemetry and, if not, flag a possible regime shift or slow-loop refresh.

Much of the existing work represents O-RAN relationships using graph-based models, such as conflict graphs, GNNs, graph convolutional networks, and hypergraphs \cite{zolghadr_conflicts,pacifista,comix,alshami_gcn,henna_hypergraph,bermudez_gnn_mobility}.
These graph representations are useful for visualization, link prediction, conflict labeling, and structural learning.
However, these methods are designed for periodic or offline analysis.
PACIFISTA reports conflict-evalution times on the order 20 seconds \cite{pacifista}.
GNN-based reconstruction requires training over hundreds of epochs \cite{zolghadr_conflicts}.
These timescales are poorly matched to tracking parameter--KPI shifts in dynamic tactical contexts.

In this paper, we use a Boolean representation of parameter--KPI relationships to enable dynamic runtime monitoring within a fast/slow-loop workflow.
A slower \textit{learning loop} learns a bolleanization model for detecting significant parameter-activit and KPI response events \cite{slow_loop_paper}.
The \textit{runtime loop} applies this model to streaming telemetry to maintain and update parameter--KPI dependencies.
This separation keeps the heavier continuous-data learning problem in the slow loop and leaves the fast loop with a compact algebraic tracking problem.

The main contributions of this paper are as follows:
\begin{enumerate}
	\item We frame runtime parameter--KPI dependency tracking as an AI-RAN conflict-monitoring primitive for detecting stale dependency assumptions in evolving control environments, including contested tactical settings where shifts may be abrupt and adversarially induced.

	\item We introduce a Boolean-algebraic dependency representation in which checking whether recent observations match the current dependency model reduces to a Boolean matrix equation.

	\item We define a fast/slow-loop workflow in which a slower learning stage provides Boolean event indicators, while the runtime loop maintains an interpretable dependency matrix online.

	\item We develop and evaluate a sliding-window reuse-and-recompute inference procedure, quantifying tracking accuracy, false alarms, detection delay, and per-event processing cost on controlled Boolean event streams.
\end{enumerate}

\section{Methods}

This section defines the runtime telemetry abstraction, Boolean dependency model, sliding-window inference rule, controlled Boolean event generator, and tracking metrics.

\subsection{Runtime Telemetry and Boolean Event Model}

Let \(p(c) \in \mathbb{R}^{n_P}\), \(k(c) \in \mathbb{R}^{n_K}\) denote real-valued control-parameter and KPI telemetry at observation index \(c\).
Here \(n_P\) is the number of control parameters and \(n_K\) is the number of KPIs.
At runtime, each observation is converted into Boolean event indicators using a booleanization model learned by a slower stage \cite{slow_loop_paper}:
\[
\mathcal{Q}_{\theta} : (p(c),k(c),\mathcal{H}_c) \mapsto
\left(b_P(c),b_K(c)\right),
\]
where \(\mathcal{H}_c\) denotes any recent history or baseline information used by the converter, \(b_P(c) \in \{0,1\}^{n_P}\), and \(b_K(c) \in \{0,1\}^{n_K}\).
The entry \(b_P[i](c)=1\) indicates a significant activity event in parameter \(P_i\); \(b_K[j](c)=1\) indicates a significant response in KPI \(K_j\).
This paper focuses on the dependency-tracking step after this conversion; learning the booleanization model is addressed as part of the slow loop \cite{slow_loop_paper}.

\subsection{Boolean Dependency Representation}

\begin{figure}[t]
	\centering
	\resizebox{\columnwidth}{!}{%
	\begin{tikzpicture}[
		font=\small,
		title/.style={font=\bfseries},
		cell/.style={draw=black, minimum width=8mm, minimum height=7mm, inner sep=0pt},
		one/.style={cell, fill=green!20},
		zero/.style={cell, fill=white},
		vec/.style={draw=black, minimum width=6mm, minimum height=7mm, inner sep=0pt},
		pactive/.style={circle, draw=blue!70!black, fill=blue!18, thick, minimum size=8mm, inner sep=0pt},
		inactive/.style={circle, draw=black!65, fill=gray!12, thick, minimum size=8mm, inner sep=0pt},
		kinactive/.style={circle, draw=yellow!50!black, fill=yellow!25, thick, minimum size=8mm, inner sep=0pt},
		kactive/.style={circle, draw=orange!80!black, fill=orange!32, thick, minimum size=8mm, inner sep=0pt},
		kmismatch/.style={circle, draw=red, fill=yellow!25, very thick, minimum size=8mm, inner sep=0pt},
		legendtext/.style={font=\small, align=left, anchor=west}
		]
		
		% =========================================================
		% Panel boundaries / separators
		% =========================================================
		\draw[dashed, gray!80] (3.20,1.35) -- (3.20,-3.25);
		\draw[dashed, gray!80] (8.45,1.35) -- (8.45,-3.25);
		
		% =========================================================
		% (a) Matrix L
		% =========================================================
		\node[title, anchor=west] at (-0.10,1.15) {(a) Matrix $L$};
		
		% Column labels
		\node at (0.95,0.20) {$P_1$};
		\node at (1.75,0.20) {$P_2$};
		\node at (2.55,0.20) {$P_3$};
		
		% Row labels
		\node at (0.25,-0.45) {$K_1$};
		\node at (0.25,-1.15) {$K_2$};
		
		% Matrix entries
		\node[one]  at (0.95,-0.45) {$1$};
		\node[zero] at (1.75,-0.45) {$0$};
		\node[one]  at (2.55,-0.45) {$1$};
		
		\node[zero] at (0.95,-1.15) {$0$};
		\node[one]  at (1.75,-1.15) {$1$};
		\node[one]  at (2.55,-1.15) {$1$};
		
		% =========================================================
		% (b) Single-step Boolean check
		% =========================================================
		\node[title, anchor=west] at (3.65,1.15) {(b) Boolean check};
		
		% Formula labels, spaced apart explicitly
		\node at (4.40,0.30) {$p(t)$};
		\node at (6,0.30) {$\widehat{k}(t)$};
		\node at (7.73,0.30) {$k(t)$};
		
		% p(t) vector
		\node at (3.80,-0.45) {$P_1$};
		\node at (3.80,-1.15) {$P_2$};
		\node at (3.80,-1.85) {$P_3$};
		
		\node[vec, fill=blue!12] at (4.35,-0.45) {$1$};
		\node[vec, fill=white]   at (4.35,-1.15) {$0$};
		\node[vec, fill=blue!12] at (4.35,-1.85) {$1$};
		
		\node[align=center] at (4.35,-2.55) {Observed};
		
		% Arrow
		\node[font=\large] at (5.05,-1.15) {$\rightarrow$};
		
		% Predicted vector
		\node at (5.5,-0.75) {$K_1$};
		\node at (5.5,-1.45) {$K_2$};
		
		\node[vec, fill=orange!25] at (6.05,-0.75) {$1$};
		\node[vec, fill=orange!25, draw=red, thick] at (6.05,-1.45) {$1$};
		
		\node[align=center] at (6.05,-2.15) {Predicted};
		
		% Not equal
		\node[font=\Large] at (6.75,-1.10) {$\neq$};
		
		% Observed vector
		\node at (7.2,-0.75) {$K_1$};
		\node at (7.2,-1.45) {$K_2$};
		
		\node[vec, fill=orange!25] at (7.75,-0.75) {$1$};
		\node[vec, fill=yellow!25, draw=red, thick] at (7.75,-1.45) {$0$};
		
		\node[align=center] at (7.75,-2.15) {Observed};
		
		\node[red] at (6.75,-2.65) {$\widehat{k}(t)\neq k(t)$};
		
		% =========================================================
		% (c) Graph view
		% =========================================================
		\node[title, anchor=west] at (8.75,1.15) {(c) Graph view };
		
		% Node coordinates.
		% P2 and K2 have the same y-coordinate, so P2--K2 is one straight horizontal segment.
		\coordinate (P1) at (9.35,0.35);
		\coordinate (P2) at (9.35,-0.75);
		\coordinate (P3) at (9.35,-1.85);
		
		\coordinate (K1) at (10.85,0.05);
		\coordinate (K2) at (10.85,-1.1);
		
		% Edges: all straight single segments
		\draw[thick] (P1) -- (K1);
		\draw[thick] (P3) -- (K1);
		\draw[thick] (P2) -- (K2);
		\draw[thick] (P3) -- (K2);
		
		% Nodes
		\node[pactive]  at (P1) {$P_1$};
		\node[inactive] at (P2) {$P_2$};
		\node[pactive]  at (P3) {$P_3$};
		
		\node[kactive]   at (K1) {$K_1$};
		\node[kmismatch] at (K2) {$K_2$};
		
		% Graph annotations, spaced to avoid overlap
		\node[orange!85!black, align=left, anchor=west] at (11.35,0.15) {~};
		
		\node[red, draw=red, regular polygon, regular polygon sides=3,
		minimum size=7mm, inner sep=0pt] at (11.4,-1.5) {};
		\node[red, font=\bfseries] at (11.4,-1.5) {!};
		
		%		\node[red, align=left, anchor=west] at (11.35,-1.3)
		%		{\textbf{mismatch}};
		
		\node at (9.35,-2.6) {Parameters};
		\node at (10.85,-2.6) {KPIs};
		
		% =========================================================
		% Legend: two rows to avoid crowding
		% =========================================================
		\begin{scope}[shift={(0,-4.05)}]
			\node[pactive, minimum size=5mm] at (0.15,0.4) {};
			\node[legendtext] at (0.40,0.4) {$P_i=1$ };
			
			\node[inactive, minimum size=5mm] at (2.55,0.4) {};
			\node[legendtext] at (2.80,0.4) {$P_i=0$ };
			
			\node[kactive, minimum size=5mm] at (4.95,0.4) {};
			\node[legendtext] at (5.20,0.4) {$K_i=1$ };
			
			\node[kinactive, minimum size=5mm] at (7.60,0.4) {};
			\node[legendtext] at (7.85,0.4) {$K_i=0$ };
			
			\node[kmismatch, minimum size=5mm] at (10.25,0.4) {};
			\node[legendtext] at (10.50,0.4) {Mismatch};
		\end{scope}
		
	\end{tikzpicture}%
	}
	
	\caption{Single-step Boolean dependency consistency check.
		(a) A binary matrix \(L\) encodes parameter--KPI dependencies.
		(b) The Boolean prediction \(\widehat{k}(t)=L\odot_{\mathsf B}p(t)\) is compared with the observed KPI-event vector \(k(t)\).
		(c) The same mismatch is shown on the corresponding dependency graph. Persistent mismatches over a window may indicate a change in \(L\).}
	\label{fig:single_step_boolean_check}
\end{figure}
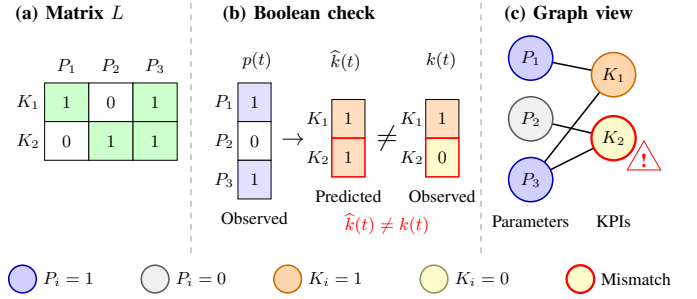

The active parameter--KPI dependency structure at observation \(c\) is represented by a latent Boolean matrix \(L^{\!*}(c) \in \{0,1\}^{n_K \times n_P}\), where \(L^{\!*}(c)[j,i]=1\) means that activity in parameter \(P_i\) can influence KPI \(K_j\) in the current operating regime.
In the ideal noiseless case:
\begin{equation}
	b_K(c) = L^{\!*}(c) \odot_{\mathsf{B}} b_P(c),
	\label{eq:ideal-boolean-model}
\end{equation}
where \(\odot_{\mathsf{B}}\) denotes Boolean matrix multiplication: \(\left(L^{\!*}(c) \odot_{\mathsf{B}} b_P(c)\right)_j = \bigvee_{i=1}^{n_P} \left( L^{\!*}(c)[j,i] \wedge b_P[i](c) \right)\).
Given a candidate matrix \(L\), the corresponding predicted KPI-event vector is \(\widehat{b}_K(c) = L \odot_{\mathsf{B}} b_P(c)\).
A single observation is explained by \(L\) when
\begin{equation}
	\widehat{b}_K(c)=b_K(c).\label{eq:eq_fails}
\end{equation}
Otherwise, the observation produces a mismatch, see \Cref{fig:single_step_boolean_check}.
The runtime objective is to maintain an estimate \(\widehat{L}(c) \in \{0,1\}^{n_K \times n_P}\) that tracks \(L^{\!*}(c)\) without treating every single mismatch as a structural change.

For a sliding window of length \(W\), define \(B_P^{(c,W)} = \left[ b_P(c-W+1),\ldots,b_P(c) \right] \in \{0,1\}^{n_P \times W}\) and \(B_K^{(c,W)} = \left[ b_K(c-W+1),\ldots,b_K(c) \right] \in \{0,1\}^{n_K \times W}\) as the stacked Boolean parameter and KPI (respectively) event matrices over the window.
A candidate dependency matrix \(L\) is consistent with the window if
\begin{equation}
	B_K^{(c,W)}
	=
	L \odot_{\mathsf{B}} B_P^{(c,W)}.
	\label{eq:window-consistency}
\end{equation}

\subsection{Row-Wise Boolean Inference}

The window-level consistency condition in \cref{eq:window-consistency} decomposes by KPI row.
For each row \(\ell_j \in \{0,1\}^{n_P}\), the row-inference problem is:
\begin{equation}
	\widehat{\ell}_j =
	\argmin_{\ell \in \{0,1\}^{n_P}}
	\left\|
	B_{K,j}^{(c,W)} - \ell \odot_{\mathsf{B}} B_P^{(c,W)}
	\right\|_0,
	\label{eq:row-inference}
\end{equation}
where \(\|\cdot\|_0\) denotes the number of nonzero entries.
Ties are broken by minimum Hamming distance from the current row estimate.
Each row of \(\widehat{L}\) can be inferred independently and in parallel, so wall-clock time does not grow with \(n_K\). For the small matrices studied here, we use exhaustive search over \(2^{n_P}\) candidates; larger deployments can apply standard 0--1 optimization methods \cite{nemhauser_wolsey_integer}.

\subsection{Sliding-Window Runtime Inference}

\cref{alg:sliding-window} maintains a sliding window of the \(W\) most recent Boolean event observations.
At each new observation index \(c\), the loop receives real-valued telemetry, converts it into Boolean event vectors, updates the window, and tests whether the current dependency estimate remains consistent with the recent window.
If the estimate is consistent, the current matrix is reused.
If not, the matrix is recomputed from the current window using the row-wise inference rule in \cref{eq:row-inference}.

\begin{algorithm}[t]
	\caption{Sliding-window dependency tracking}
	\label{alg:sliding-window}
	\DontPrintSemicolon
	\KwIn{Telemetry streams, $\mathcal{Q}_{\theta}$, window size $W$}
	Initialize window buffers and current estimate $\widehat{L}_{cur}$\;
	\For{$c=1,\ldots,T$}{
		$(b_P(c),b_K(c)) \leftarrow \mathcal{Q}_{\theta}(p(c),k(c),\mathcal{H}_c)$\;
		Update $B_P^{(c,W)}$ and $B_K^{(c,W)}$\;
		\If{window is full \textbf{and} ($\widehat{L}_{cur}$ is empty \textbf{or} \cref{eq:eq_fails} fails)}{
			Recompute $\widehat{L}_{cur}$ using \cref{eq:row-inference}\;
		}
		Set $\widehat{L}(c)\leftarrow \widehat{L}_{cur}$\;
	}
\end{algorithm}

This reuse-and-recompute rule is designed to avoid unnecessary inference.
Recomputation is triggered only when recent observations are inconsistent with the current matrix.
The window size \(W\) controls the main stability--adaptability tradeoff.
A small \(W\) can react quickly to structural changes but may be underdetermined and sensitive to isolated booleanization errors.
A large \(W\) provides more evidence for each inferred dependency matrix but can mix observations from multiple dependency regimes after a change.

\subsection{Dependency-Tracking Data Generator}

The experimental evaluation isolates the runtime tracking layer.
Therefore, we directly generate controlled Boolean event streams with known ground-truth dependency changes.
This allows us to study the behavior of the sliding-window tracker independently of errors introduced by the continuous-to-Boolean conversion stage.

An initial sparse matrix \(L^{\!*}(1) \in \{0,1\}^{n_K \times n_P}\) is perturbed every \(D\) observations to produce a piecewise-constant ground-truth sequence \(L^{\!*}(1),\ldots,L^{\!*}(T)\).
Larger values of \(D\) correspond to slower structural change, while smaller values of \(D\) create a more dynamic tracking problem.
Parameter-event vectors \(b_P(c)\) are sampled at each step.
The corresponding noiseless KPI-response vector is \(b_K(c) = L^{\!*}(c) \odot_{\mathsf{B}} b_P(c)\).
Boolean noise is added by independently flipping \(b_K(c)\) entries with probability \(\epsilon\).

\subsection{Evaluation Metrics}

We evaluate the inferred sequence $\widehat{L}(c)$ against the ground-truth sequence $L^{\!*}(c)$ using both edge-level and change-detection metrics.
For edge-level tracking, each entry of the dependency matrix is treated as a binary edge decision between one parameter and one KPI.
We flatten these matrix entries across all evaluated observations and compute standard binary precision, recall, and F1 score~\cite{powers2011evaluation}.

We also report matrix-level structural error using the Hamming distance \(d_H(\widehat{L}(c),L^{\!*}(c)) = \sum_{j=1}^{n_K}\sum_{i=1}^{n_P} \mathbf{1}\{\widehat{L}(c)[j,i]\neq L^{\!*}(c)[j,i]\}\).

For change detection, a true change occurs when $L^{\!*}(c) \neq L^{\!*}(c-1)$, and a detected change occurs when $\widehat{L}(c) \neq \widehat{L}(c-1)$.
Let $\tau$ denote the change-matching tolerance in observations.
A detected change at $\widehat{c}$ is matched to a true change at $c_0$ only if
\[
c_0 \leq \widehat{c} \leq c_0+\tau .
\]
The first such detected change is counted as the detection for that true change, with delay $\widehat{c}-c_0$.
Additional inferred changes during the same post-change adaptation period are counted separately as adaptation updates rather than as additional successes.
Detected changes outside any allowed post-change region are counted as false alarms.
We report change recall, change precision, change F1, false-alarm fraction, and mean/median detection delay in observations.
This avoids over-crediting unstable inference that changes frequently: frequent updates may increase raw recall, but they also increase false alarms and adaptation updates.

\section{Experimental Results and Discussion}

This section evaluates the runtime dependency-tracking loop under controlled dynamic conditions.
The experiments use Boolean event streams with known dependency matrices and planted structural changes, so the inferred dependency sequence can be compared against a known reference.
All reported plots use the representative moderate-noise setting \(\epsilon=0.02\), so the discussion focuses on the interaction between structural-change rate and sliding-window tracking.
A broader sweep over event-noise levels is left for future evaluation.

\subsection{Experimental Setup}

Unless otherwise stated, the experiments use $n_P=5$ parameters, $n_K=4$ KPIs, and sparse dependency matrices, generated across multiple random seeds, with at most three active parameter parents per KPI row.
We vary the dependency-change interval
$D \in \{32,64,75,100,150,300\}$,
and the sliding-window size
$W \in \{4,8,16,32\}$.
Each configuration is repeated across multiple random seeds.
For each run, we generate a dynamic Boolean stream, infer the dependency sequence with Algorithm~\ref{alg:sliding-window}, and compute the metrics described above.

For the change-detection metrics, we set the matching tolerance to $\tau=W$ observations after each true dependency change, and use the same $W$-observation interval as the adaptation grace period for counting repeated post-change updates.

Two diagnostic considerations are especially important when interpreting the results.
First, the sliding-window size \(W\) should be interpreted relative to the dependency-change interval \(D\). 
Larger windows provide more evidence for each inferred dependency matrix, but they are also more likely to include observations from before and after a dependency change, weakening the approximate stationarity assumption inside the window.
Second, the false-alarm and adaptation-update metrics are reported together with change recall to distinguish genuine change detection from unstable repeated updating.

\subsection{Runtime Cost of Reuse and Recomputation}

\begin{figure}[t]
\centering
\maybeincludegraphics[width=0.99\columnwidth]{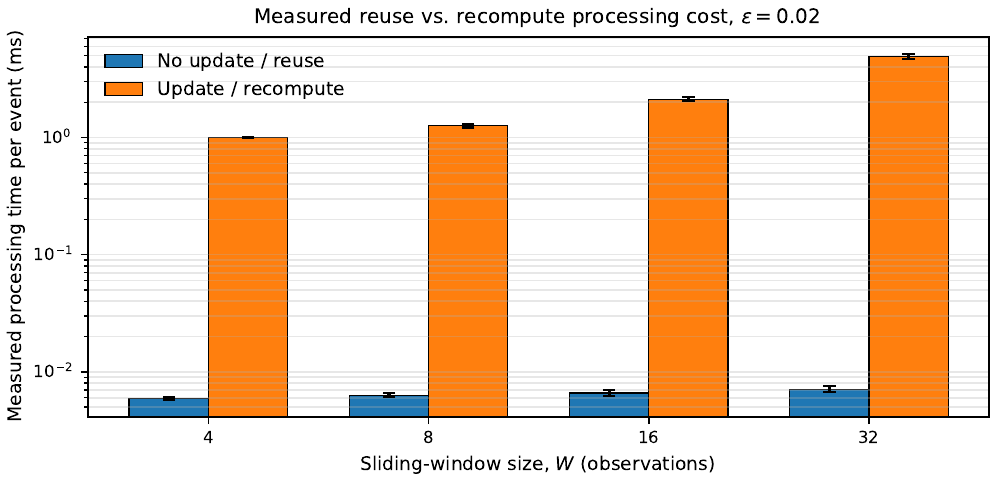}
\caption{Measured per-event processing cost for the no-update/reuse and update/recompute paths at \(\epsilon=0.02\). Error bars show bootstrapped 10th--90th percentile ranges.}
\label{fig:reuse-update-timing}
\end{figure}

Fig.~\ref{fig:reuse-update-timing} reports the measured processing cost of the two runtime paths in Algorithm~\ref{alg:sliding-window}.
The timing results show that reuse events are essentially independent of the sliding-window size $W$.
This is expected because the reuse path only evaluates the current matrix on the newest parameter-event column.
Across all tested window sizes, the measured reuse cost remains well below $0.01$ ms per event.
By contrast, recomputation is substantially more expensive and increases with $W$, since the update path solves the row-wise inference problem over the full window.
Thus, the measured runtime behavior supports the intended reuse-and-recompute design: most observations can be handled by a very fast single-column consistency check, while the more expensive window-level inference step is reserved for observations that are inconsistent with the current dependency estimate.

Existing O-RAN conflict-management methods address broader graph-reconstruction or conflict-evaluation tasks that are heavier than the per-event consistency check studied here.
For example, PACIFISTA profiles applications in sandbox environments and reports that its prototype conflict-evaluation procedure takes approximately 20 s~\cite{pacifista}.
GNN-based conflict-graph reconstruction methods require training over hundreds of epochs~\cite{zolghadr_conflicts} to recover hidden xApp--parameter--KPI relationships.
These methods are complementary rather than direct baselines: they reconstruct or evaluate conflicts, whereas the proposed Boolean tracker assumes Boolean events are available and tests whether the current dependency support remains consistent with each new observation.
Thus, the timing value of the proposed method is as an always-on guard that can trigger slower graph-reconstruction or conflict-evaluation procedures only when recent telemetry suggests a stale dependency model.

\subsection{Tracking Accuracy}

\begin{figure}[t]
\centering
\subfigure[Edge-level F1]{%
\includegraphics[width=0.48\columnwidth]{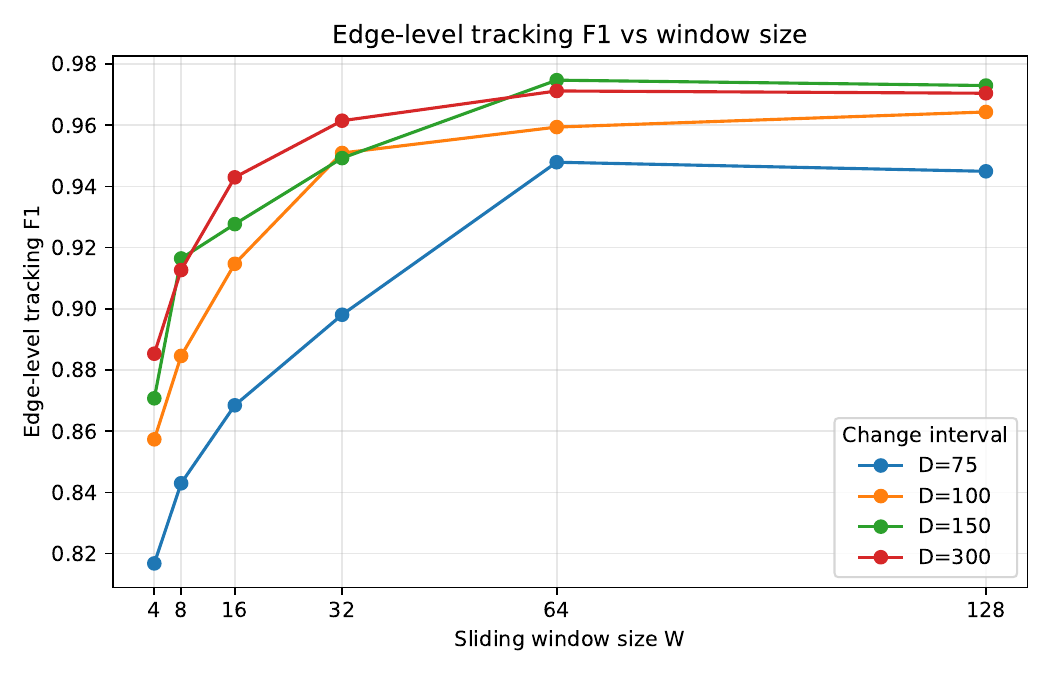}
}
\hfill
\subfigure[Mean Hamming distance]{%
\includegraphics[width=0.48\columnwidth]{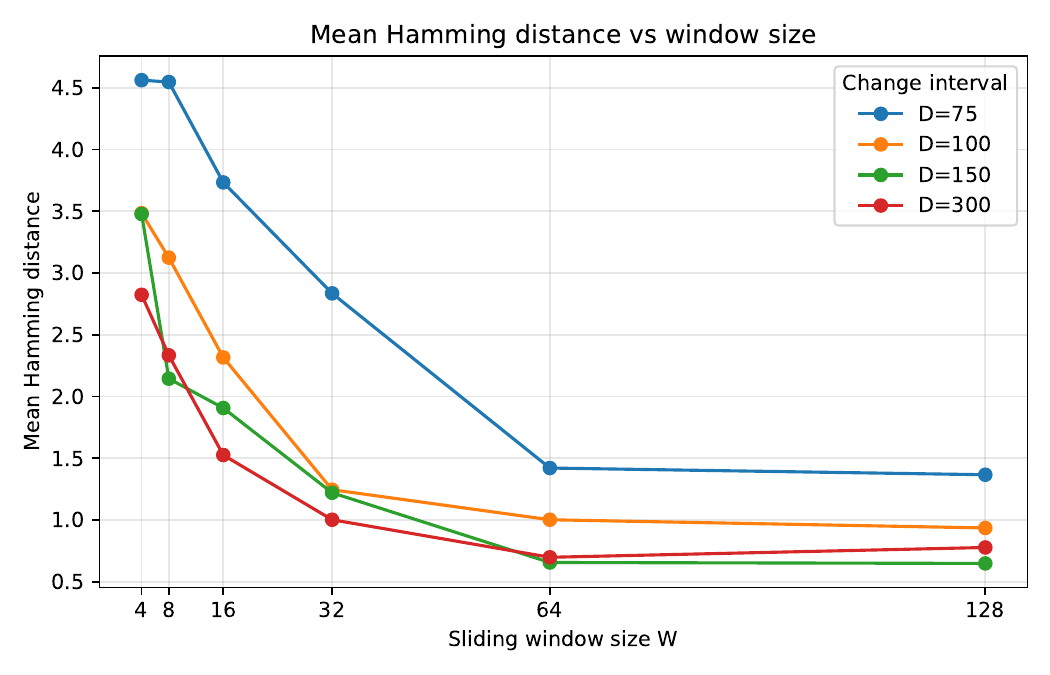}
}
\caption{Tracking accuracy and structural error versus sliding-window size at \(\epsilon=0.02\): (a) edge-level F1 and (b) mean Hamming distance.}
\label{fig:tracking-accuracy}
\end{figure}

Fig.~\ref{fig:tracking-accuracy} reports both edge-level tracking accuracy and matrix-level structural error.
Panel~(a) shows edge-level F1, where each parameter--KPI matrix entry is treated as one binary dependency prediction.
Panel~(b) shows mean Hamming distance, which measures how many entries of the inferred matrix are incorrect on average.

Together, these metrics distinguish average edge recovery from the severity of matrix-level errors.
A high edge-level F1 indicates that most dependencies are recovered correctly, while a low Hamming distance indicates that the inferred matrix is usually only a few entries away from the true matrix.
The results show the same stability--adaptability trend observed throughout the paper: moderate windows provide enough evidence for inference, while overly large windows can mix observations from multiple dependency regimes after a structural change.

\subsection{Change Detection, False Alarms, and Delay}

\begin{figure*}[t]
\centering
\subfigure[Change recall, $\epsilon=0.02$]{%
\maybeincludegraphics[width=0.31\linewidth]{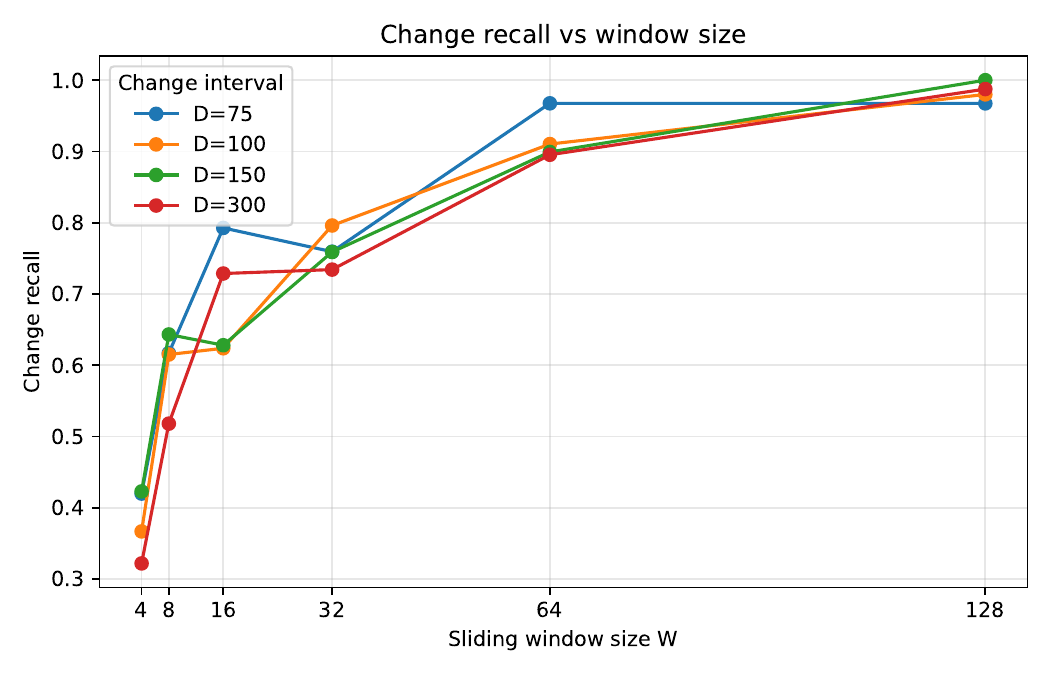}}
\hfill
\subfigure[False-alarm fraction, $\epsilon=0.02$]{%
\maybeincludegraphics[width=0.31\linewidth]{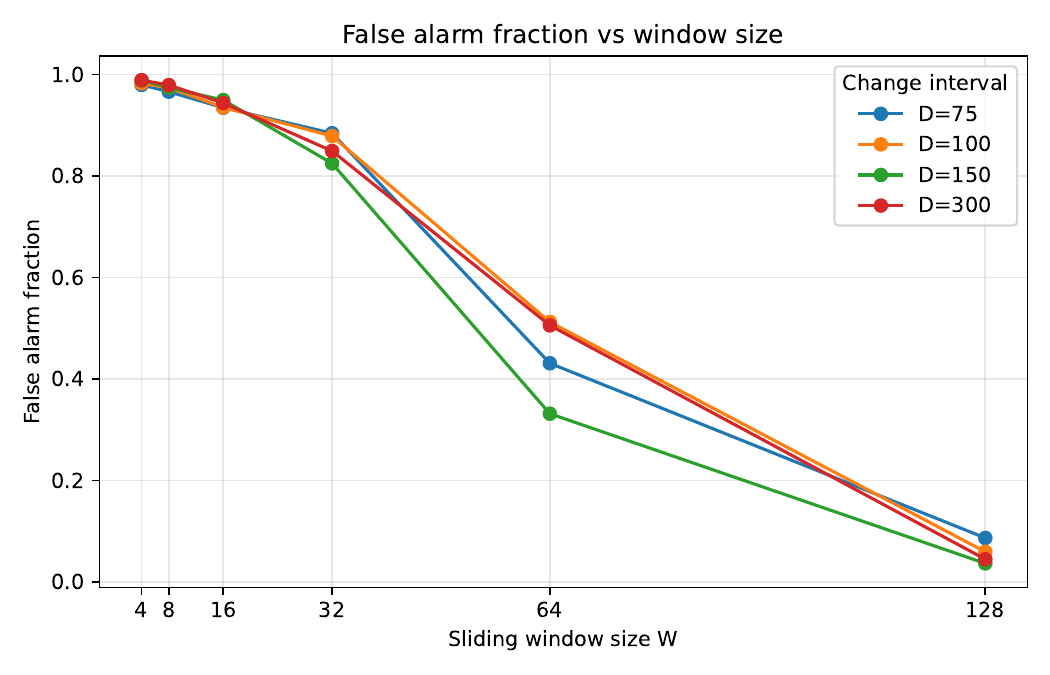}}
\hfill
\subfigure[Mean detection delay, $\epsilon=0.02$]{%
\maybeincludegraphics[width=0.31\linewidth]{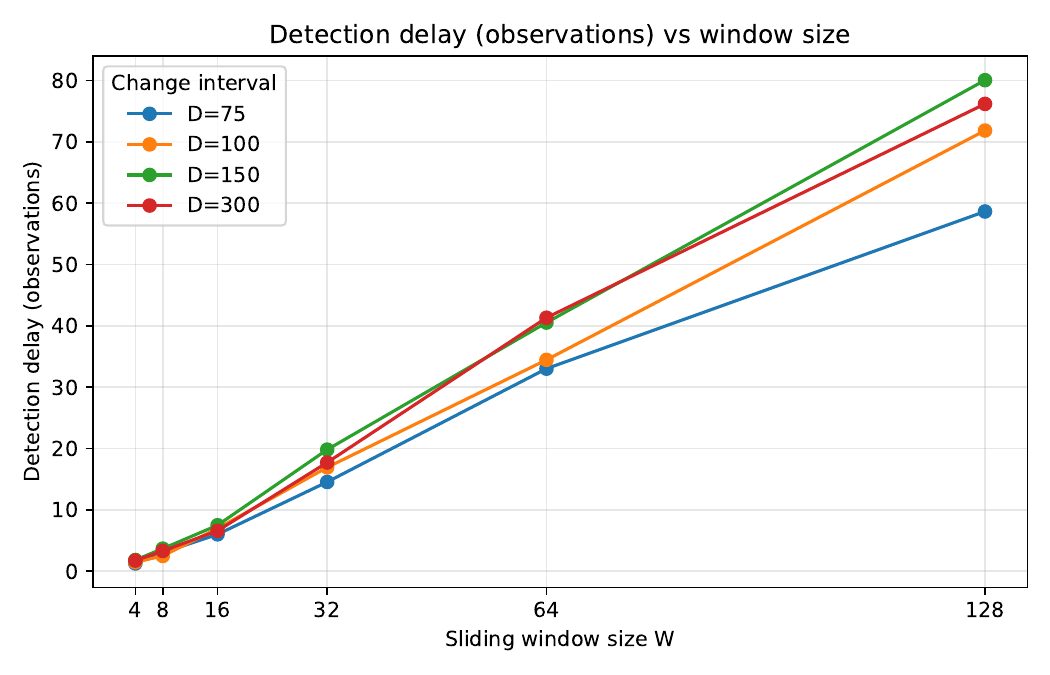}}
\caption{Change-detection tradeoff at \(\epsilon=0.02\). Larger windows generally reduce false alarms but increase detection delay.}
\label{fig:change-detection-tradeoff}
\end{figure*}

Fig.~\ref{fig:change-detection-tradeoff} summarizes the change-detection behavior for the representative moderate-noise condition $\epsilon=0.02$.
Panel (a) reports change recall, panel (b) reports the false-alarm fraction, and panel (c) reports mean detection delay.
These metrics must be interpreted jointly.
A method can obtain high change recall simply by changing its inferred matrix very often, but such behavior is not operationally useful if it also produces many false alarms.
The updated evaluation therefore matches at most one detected change to each true dependency change and counts extra updates during the adaptation region separately.

In the representative condition shown, larger windows tend to reduce false alarms because the inferred matrix is based on more evidence.
However, larger windows also increase detection delay.
This is the central dynamicity result of the paper: window size controls a tradeoff between stable tracking and fast responsiveness.
Small windows react quickly but may update too often; large windows are more stable but react later and can mix observations from before and after a dependency change.

Overall, these results show that the Boolean tracker is most useful when the dependency structure remains stable long enough for the window to collect informative observations. Larger \(W\) improves stability and reduces false alarms, but increases detection delay and can mix dependency regimes. In an AI-RAN/O-RAN setting, the tracker should therefore be interpreted as an explainable runtime monitor that flags possible dependency shifts or model-refresh needs, rather than as an autonomous mitigation mechanism.

\subsection{Limitations}

The present study is intentionally scoped.
The experiments use synthetic Boolean streams with planted dependency changes.
The Boolean model captures only the support of parameter--KPI influence, not direction, magnitude, timing, causal strength, or multi-parameter interaction effects.
Noise is modeled as independent Boolean flips in KPI events, and the experiments do not yet include a continuous telemetry source or a full RAN simulator. The reported plots use a representative moderate Boolean KPI-noise setting, \(\epsilon=0.02\). 
A broader sweep over booleanization error rates is left for future work.
The row-wise exhaustive inference procedure is appropriate for the small matrices studied here, but larger deployments will require constrained or approximate search.
A direct empirical comparison with graph-reconstruction or full conflict-evaluation frameworks is left for future work, since those methods solve broader tasks than the per-event dependency-consistency check isolated here.
These limitations do not invalidate the tracking formulation, but they define the next steps required before the approach can be evaluated as a practical AI-RAN monitoring component.

\section{Conclusion}

This paper studied interpretable runtime dependency tracking as one step toward explainable monitoring for AI-RAN control loops.
Assuming that continuous telemetry has already been converted into Boolean parameter-activity and KPI-response events, we formulated runtime dependency maintenance as a streaming tracking problem over a time-varying Boolean parameter--KPI matrix.
We proposed a lightweight reuse-and-recompute sliding-window inference rule and evaluated tracking quality under controlled dependency changes.

The results emphasize the stability--adaptability tradeoff.
Longer dependency-change intervals make tracking easier because the active structure remains stable for more observations.
Larger windows provide more stable inference and reduce false alarms, but they also increase detection delay and can mix multiple dependency regimes.
These trends indicate that runtime Boolean dependency tracking is most useful when window size is chosen relative to the expected rate of structural change.

The tracker is therefore best viewed as a compact runtime monitoring component rather than a complete AI-RAN conflict-mitigation system.
Future work will connect it to continuous-data learning, evaluate richer AI-RAN-inspired traces or simulators, compare against streaming baselines, and study how detected dependency shifts support conflict diagnosis, mitigation, and policy adaptation.

\bibliographystyle{IEEEtran}
\bibliography{refs}

\end{document}